\documentclass[letterpaper, 10 pt, conference]{ieeeconf}  

\IEEEoverridecommandlockouts                              

\usepackage{cite}
\usepackage{amsmath,amssymb,amsfonts}
\usepackage{algorithmic}
\usepackage{graphicx}
\usepackage{textcomp}
\usepackage{xcolor}
\usepackage{url}

\title{CLMN: Concept based Language Models via Neural Symbolic Reasoning\\}
\author{Yibo Yang$^{1}$
\thanks{$^{1}$Yibo Yang is with the College of Science, The Hong Kong University of Science and Technology, Hong Kong {\tt\small yyanggh@connect.ust.hk}}}

\begin{document}

\maketitle
\thispagestyle{empty}
\pagestyle{empty}

\begin{abstract}

Deep learning’s remarkable performance in natural language processing (NLP) faces critical interpretability challenges, particularly in high-stakes domains like healthcare and finance where model transparency is essential. While concept bottleneck models (CBMs) have enhanced interpretability in computer vision by linking predictions to human-understandable concepts, their adaptation to NLP remains understudied with persistent limitations. Existing approaches either enforce rigid binary concept activations that degrade textual representation quality or obscure semantic interpretability through latent concept embeddings, while failing to capture dynamic concept interactions crucial for understanding linguistic nuances like negation or contextual modification. This paper proposes the \underline{C}oncept \underline{L}anguage \underline{M}odel \underline{N}etwork (\underline{CLMN}), a novel neural-symbolic framework that reconciles performance and interpretability through continuous concept embeddings enhanced by fuzzy logic-based reasoning. CLMN addresses the information loss in traditional CBMs by projecting concepts into an interpretable embedding space while preserving human-readable semantics, and introduces adaptive concept interaction modeling through learnable neural-symbolic rules that explicitly represent how concepts influence each other and final predictions. By supplementing original text features with concept-aware representations and enabling automatic derivation of interpretable logic rules, our framework achieves superior performance on multiple NLP benchmarks while providing transparent explanations. Extensive experiments across various pre-trained language models and datasets demonstrate that CLMN outperforms existing concept-based methods in both accuracy and explanation quality, establishing a new paradigm for developing high-performance yet interpretable NLP systems through synergistic integration of neural representations and symbolic reasoning in a unified concept space. Code and data are available at \url{https://github.com/MichaelYang-lyx/CLMN}.
\end{abstract}


\section{Introduction}

Deep learning has seen widespread application due to its outstanding performance in fields like image recognition \cite{bansal2023transfer} and natural language processing \cite{min2023recent}. However, the complex neural network structures and numerous parameters make its decision-making process difficult to interpret, often earning it the label of a "black box" \cite{hassija2024interpreting}. This characteristic limits the use of deep learning in areas where high interpretability is essential, such as healthcare \cite{elshawi2021interpretability}, finance \cite{lin2022model}, law \cite{wu2022towards}, and autonomous driving \cite{shao2023safety}.

In the healthcare sector, the opacity of deep learning models used for diagnosis and treatment decisions can lead to skepticism among doctors and patients \cite{quinn2022three}. For instance, when a model recommends a specific treatment plan, doctors need to understand the underlying reasons and logic to ensure its reliability and effectiveness \cite{singh2020explainable}. In finance, the use of deep learning models in risk assessment and credit scoring can raise regulatory and legal issues. Financial institutions and regulatory bodies need to be able to explain and verify the decision-making processes of these models to avoid potential compliance risks \cite{bussmann2021explainable}. 

Clearly, interpretability is crucial in deep learning, as it enhances user trust in model decisions and ensures safety and compliance in critical domains. Against this backdrop,  concept bottleneck model (CBM) has emerged as a significant research direction. However, there has been little research on CBM in the field of NLP, with most studies focusing on computer vision (CV). It wasn't until 2024 that the paper \textit{Interpreting Pretrained Language Models via Concept Bottlenecks}~\cite{tan2024interpreting} first attempted to apply CBM to pretrained language models (PLM), achieving promising results but leaving room for improvement.

The fundamental challenge lies in the inherent conflict between performance and interpretability when adapting CBM to NLP tasks. Traditional CBM enforces hard concept interventions through binary concept activation, which inevitably causes information loss in text representations. While Concept Embedding Models (CEM) \cite{zarlenga2022concept} alleviate this issue by projecting concepts into continuous embedding space, their latent space operations obscure human-understandable semantics. Furthermore, existing approaches fail to capture the dynamic interactions between concepts – for instance, how negation concepts like “not severe” modify disease severity concepts in medical texts. This limitation stems from their reliance on rigid concept labeling rather than learning conceptual relationships from data.

\begin{figure*}[htb]
\centering 
\fbox{\includegraphics[width=0.98\linewidth]{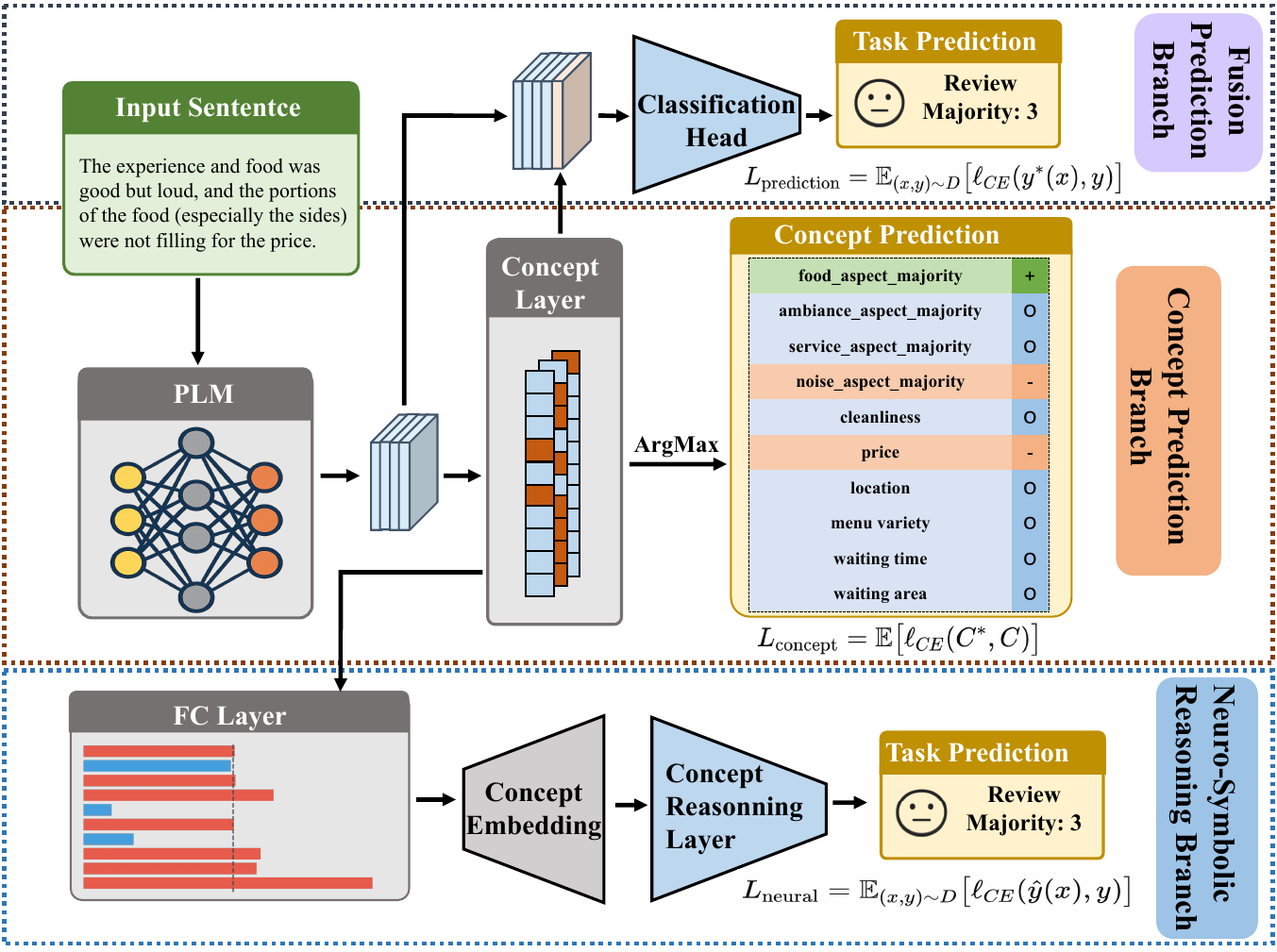}}
\caption{This flowchart details a text - classification framework. An input sentence is input into a PLM. The Concept Layer predicts aspects like food. Its results, combined with PLM's, yield a sentiment score. Concepts also flow to the Concept Reasoning Layer, constraining internal model explanations.}
\label{fig:framework}
\end{figure*}

In response, we considered CEM, which improved model performance but compromised interpretability. To address this, we further incorporated fuzzy logic for interpretable neural-symbolic concept reasoning \cite{keberneuro,zhang2024neuro}. This approach not only enhanced model performance but also improved interpretability. Through this method, we achieved concept explanations and generated logical rules composed of concepts, clarifying the relationship between predictions and concepts. The overall framework is depicted in Figure~\ref{fig:framework}.

Our contributions are as follows:
\begin{itemize}
    \item We propose a novel neural-symbolic interpretable framework for the NLP field: Concept Language Model Network (CLMN). CLMN effectively addresses longstanding issues such as the accuracy drop in concept-based approaches and the inability of early complex neural-symbolic models to reflect interactions between concepts.
    \item We utilize neural-symbolic reasoning to efficiently interpret the interactions between concepts, addressing the efficiency issues in concept-based model interactions. Additionally, using concept features as a supplement to original modality features enhances final performance accuracy.
    \item We applied our framework to various PLMs and datasets, conducting extensive experiments. The results demonstrate that our framework generates semantic explanations through neural-symbolic reasoning in a consistent concept space, offering interpretability while achieving excellent performance.
\end{itemize}
\section{Related Work}

\subsection{Pretrained Language Models}
The rapid advancement of pretrained language models (PLMs) has fundamentally transformed NLP through their capacity to capture intricate linguistic patterns. Transformer-based architectures like BERT \cite{devlin2018bert} and RoBERTa \cite{liu2019roberta} leverage bidirectional self-attention mechanisms to construct context-aware representations, with empirical studies revealing their ability to encode syntactic and semantic hierarchies across attention heads \cite{jawahar2019does}. In contrast, autoregressive models such as GPT-2 \cite{radford2019language} employ unidirectional attention to optimize generative coherence, demonstrating emergent capabilities in implicit knowledge representation through next-token prediction paradigms \cite{radford2019language}. While these transformer-based models dominate contemporary research, recurrent architectures like LSTM \cite{hochreiter1997long} maintain practical relevance by offering computationally efficient temporal modeling, particularly in scenarios with sequential output dependencies.

\subsection{Concept Bottleneck Model}

CBM \cite{koh2020concept} represents an innovative approach in deep learning for image classification and visual reasoning by introducing a concept bottleneck layer within deep neural networks. Despite its promising prospects, CBM faces several challenges. Firstly, CBM often underperforms compared to original models without the concept bottleneck layer. This performance gap arises because CBM cannot fully extract information from the raw data into the bottleneck features. Additionally, the expansion of tasks requires CBM models to balance accuracy and interpretability, whereas traditional models only focus on prediction accuracy. Secondly, CBM's effective performance hinges on extensive dataset annotations, posing a significant barrier to its widespread application. Researchers have explored various solutions to address these issues.

For instance, Oikarinen et al. \cite{oikarinen2023label} proposed a label-free CBM to overcome the limitations of traditional CBM, offering an effective alternative. Additionally, Yuksekgonul et al. \cite{yuksekgonul2022post} introduced a posterior concept bottleneck model that can be applied to various neural networks without sacrificing model performance while retaining its interpretability advantages. Furthermore, Chauhan et al. \cite{chauhan2023interactive} extended the application of CBM to interactive prediction environments by introducing interaction strategies to select annotated concepts, thereby enhancing prediction accuracy.

However, most research on CBM has predominantly focused on the CV
domain, with relatively limited exploration in NLP. While some progress has been made in applying CBMs to NLP tasks, significant gaps and opportunities for improvement remain. Among the few existing studies, the work by Zhen Tan\cite{tan2024interpreting} represents one of the early efforts to integrate CBMs into NLP tasks. However, this approach primarily inserts the concept layer prior to label prediction, which can lead to feature information loss and a decline in model accuracy. Furthermore, the relationship between the predicted concepts and the final labels remains ambiguous. As such, there is substantial potential for advancing CBM methodologies in NLP by exploring more effective strategies for incorporating concept layers while preserving feature integrity and clarifying the derivation process linking concepts to labels. This line of inquiry could significantly enhance interpretability in NLP models.

\subsection{Neural Symbolic Reasoning}

To address the lack of a clear derivation relationship between concepts and the final prediction labels in CBM, we introduce neural-symbolic reasoning techniques \cite{garcez2019neural,garcez2022neural,hitzler2022neuro,barbiero2023interpretable}. These techniques bridge the gap between symbolic reasoning, traditionally used in artificial intelligence systems, and the data-driven learning capabilities of neural networks. One method in neural-symbolic reasoning is Neuro-Symbolic Forward Reasoning (NS-FR) \cite{shindo2021neuro}. NS-FR is a hybrid approach that combines neural networks with symbolic logic reasoning. It embeds symbolic reasoning rules within neural networks, allowing for the derivation of new facts in a differentiable manner. By incorporating NS-FR, we can better interpret the relationship between concepts and prediction labels in CBM, thereby enhancing the model's interpretability and robustness. Additionally, we observe the application of neural-symbolic Visual Question Answering (VQA) \cite{yi2018neural}, which explores a method that integrates neural networks with symbolic reasoning for VQA tasks. This approach achieves a decoupling of visual understanding, language understanding, and reasoning abilities, thereby improving the model's performance in handling complex question-answering tasks. This method not only enhances the accuracy of VQA models but also provides a more transparent and interpretable decision-making process.

Inspired by these works, we explore the introduction of neural symbolic reasoning into the framework of text interpretable classification to further improve the interpretability and model performance. By incorporating the principles of NS-FR and neural-symbolic into our research on interpreting pretrained language models, we aim to achieve similar advantages within the CBM framework. This integration is expected to clarify the relationship between concepts and prediction labels, thereby increasing the practical value of the model.

\section{Methods}
\subsection{Preliminaries}

\noindent {\bf Overview of Concept Bottleneck Models (CBMs).} Utilizing the notational framework described in \cite{koh2020concept}, we delve into a classification scheme characterized by a set of defined concepts ${\mathcal{C}} = {p_1, \dots, p_k}$. The pertinent dataset is denoted by ${c}=\{p_1, \cdots, p_k\}$, where $x_i \in \mathbb{R}^d$ denotes the input features for each instance $i$ in the index set $[N]$, $y_i \in \mathbb{R}^{d_y}$ signifies the associated target labels (with $d_y$ indicating the total number of classifications), and $\mathbf{c}_i \in \mathbb{R}^k$ encapsulates the concept vectors. Here, the element $c_{ij}$ within $\mathbf{c}_i$ measures the significance of concept $p_j$ for instance $i$. The aim of CBMs is to develop two pivotal mappings: firstly, the transformation $h: \mathbb{R}^d \to \mathbb{R}^k$, which converts the feature space into concept space; secondly, the function $m: \mathbb{R}^k \to \mathbb{R}^{d_y}$, which interprets the concept space into prediction outputs. The primary goal is to ensure that the estimated concept vector $\hat{\mathbf{c}} = h(x)$ and the resulting classification $\hat{y} = m(h(x))$ closely match their actual counterparts, effectively embodying the core principles of CBMs.

\noindent {\bf Fuzzy logic rules.} Traditional Boolean logic operates on discrete truth values $\{0, 1\}$, limiting its capacity to handle uncertain concept interactions. Fuzzy logic \cite{hajek2013metamathematics} addresses this by introducing continuous truth degrees through three core operators:  

\noindent- \textbf{Conjunction} ($\land$): Implemented via t-norms, with the product operator $a \land b = ab$ ensuring differentiable gradients  

\noindent- \textbf{Disjunction} ($\lor$): Defined through t-conorms using $a \lor b = a + b - ab$  

\noindent- \textbf{Negation} ($\neg$): Standard negation $\neg a = 1 - a$  

Consider classifying "apple" using concepts $\{c_{\text{red}}, c_{\text{round}}, c_{\text{crisp}}\}$. A fuzzy rule might express:  
\begin{equation} 
y_{\text{apple}} = (\neg c_{\text{red}} \land c_{\text{round}}) \lor c_{\text{crisp}}
\end{equation}  

This captures two pathways: 1) non-red coloration with round shape, or 2) crisp texture regardless of color. For an apple with partial redness ($c_{\text{red}}=0.4$), perfect roundness ($c_{\text{round}}=1$), and moderate crispness ($c_{\text{crisp}}=0.7$):  

\begin{equation} 
y_{\text{apple}} = (0.6 \land 1) \lor 0.7 = (0.6 \times 1) \lor 0.7 = 0.6 + 0.7 - 0.42 = 0.88 
\end{equation}  

The continuous formulation enables gradient-based optimization of concept compositions while preserving logical interpretability \cite{barbiero2023interpretable}. This reconciles discrete rule-based reasoning with neural feature learning through differentiable surrogates of logical operators.

\subsection{Pretrained Language Model}

Our objective is to predict the target label $ y $ given an input text $ x $. We employ a parameterized pre-trained language model as the foundation of our approach. Initially, the input text $ x $ is fed into the model, which encodes it into a latent representation $ z $. This process can be formulated as:

\begin{equation}
x \to z \to y
\end{equation}

\noindent where $ x $ is transformed into the latent representation $ z $, which is subsequently mapped to the predicted label $ y $. The overall performance of the task is primarily determined by the accuracy of the predicted $ y $.

\subsection{Concept Layer and Concept Embedding}

To enhance the interpretability of the model and provide a more explainable prediction for the output label $ y $, we introduce the Concept Layer and Concept Embedding approach. This method explicitly integrates conceptual representations into the model, ensuring a more transparent decision-making process.

\subsubsection{Concept Layer}

The Concept Layer explicitly encodes the state probabilities for each concept into a structured vector. Given $ S $ concepts, the Concept Layer is represented as:

\begin{equation}
\mathbf{CL} = \begin{bmatrix}
p_1^{pos} & p_1^{neg} & p_1^{unk} \\
p_2^{pos} & p_2^{neg} & p_2^{unk} \\
\vdots & \vdots & \vdots \\
p_S^{pos} & p_S^{neg} & p_S^{unk}
\end{bmatrix} \in \mathbb{R}^{S \times 3}
\end{equation}

The Concept Layer vector $\mathbf{CL}$ provides a clear interpretation of concept states, which is then utilized for downstream representation learning, activation, and embedding computations.

\subsubsection{Concept Representation.}
Given an input concept feature vector, we define a set of predicted concepts $ \mathcal{C}^{*} $ that capture high-level semantic information relevant to the task. Each concept $ c_s \in \mathcal{C} $ can take binary values, indicating its presence or absence in the input. To represent the concept states, we define:

\begin{equation}
C_s = \begin{bmatrix} C_s^+ \\ C_s^- \end{bmatrix} \in \mathbb{R}^{2e}
\end{equation}

\noindent where $ C_s^+ $ represents the active (true) state of the concept, and $ C_s^- $ represents the inactive (false) state. These embeddings are computed through concept-specific transformation layers applied to the input feature vector:

\begin{equation}
C_s^+ = \sigma(W_s^+ x + b_s^+), \quad C_s^- = \sigma(W_s^- x + b_s^-)
\end{equation}

\noindent where $ W_s^+ $ and $ W_s^- $ are learnable parameters, and $ \sigma $ is a non-linear activation function.

\subsubsection{Concept Activation and Embedding Computation.}
To determine the likelihood of a concept being active, we introduce a scoring function:

\begin{equation}
p_s = \sigma(W_q C_s + b_q)
\end{equation}

\noindent where $ p_s $ represents the probability that concept $ c_s $ is active. Using this probability, we construct the final concept embedding as a weighted sum of the active and inactive states:

\begin{equation}
\hat{c}_s = p_s C_s^+ + (1 - p_s) C_s^- \in \mathbb{R}^e
\end{equation}

Stacking all concept embeddings, we obtain the complete concept embedding matrix:

\begin{equation}
\hat{C} = \text{Concat}(\hat{c}_1, \hat{c}_2, ..., \hat{c}_S) \in \mathbb{R}^{S \times e}
\end{equation}

\subsubsection{Integration with the Prediction Model.}
Instead of relying solely on concept-based predictions, we treat the concept embeddings as an augmentation of the original feature representation. The fused representation is computed as:

\begin{equation}
F^*(x) = \text{ReLU}(W_l^T [x, \hat{C}] + b_l)
\end{equation}

\noindent where $ W_l $ and $ b_l $ are learnable parameters. The final classification output is then obtained via a linear transformation:

\begin{equation}
y^*(x) = \omega^T F^*(x) + b
\end{equation}

\subsection{Neural-Symbolic Concept Reasoning}  
Given an input instance $x$ with its unified concept embedding matrix $\hat{C} \in \mathbb{R}^{|\mathcal{C}| \times e}$, where $\mathcal{C}$ denotes the concept set and $e$ the embedding dimension, we establish interpretable decision rules through differentiable logic operations. Inspired by recent advances in neuro-symbolic integration \cite{barbiero2023interpretable}, our framework employs two complementary neural operators to model concept interactions:  

\subsubsection{Concept Polarity Network.} ($\Phi_{j,s}: \mathbb{R}^e \rightarrow [0,1]$): For each target class $j$ and concept $c_s \in \mathcal{C}$, this two-layer MLP learns to quantify the directional influence of concept embeddings through sigmoid activation:  
   \begin{equation}
   I_{p,s,j} = \sigma\left(W^{(2)}_j \cdot \text{ReLU}(W^{(1)}_j \hat{c}_s + b^{(1)}_j) + b^{(2)}_j\right)
   \end{equation}
   where $I_{p,s,j} \rightarrow 1$ indicates positive correlation with class $j$, while $I_{p,s,j} \rightarrow 0$ suggests inhibitory effects.

\subsubsection{Concept Relevance Network.} ($\Psi_{j,s}: \mathbb{R}^e \rightarrow [0,1]$): This parallel network architecture with identical depth computes context-aware concept importance weights:  
   \begin{equation}
   I_{r,s,j} = \sigma\left(V^{(2)}_j \cdot \text{ReLU}(V^{(1)}_j \hat{c}_s + d^{(1)}_j) + d^{(2)}_j\right)
   \end{equation}  
   
Higher $I_{r,s,j}$ values signify stronger evidential support from concept $c_s$ for class $j$ in instance $x$.

The final prediction rule synthesizes these signals through fuzzy logic operations:  
\begin{equation}
\hat{\mathbf{y}}_j(x) = \bigwedge_{c_s \in \mathcal{C}} \left( \neg I_{p,s,j} \lor I_{r,s,j} \right) = \min_{c_s \in \mathcal{C}} \left\{ \max\left(1-I_{p,s,j}, I_{r,s,j}\right) \right\}
\label{eq:logic_rule}
\end{equation}

To optimize this neuro-symbolic layer, we minimize a cross-entropy loss over the rule-based predictions:  
\begin{equation}
\mathcal{L}_{\text{neural}} = \mathbb{E}_{(x,y)\sim\mathcal{D}} \left[ \ell_{\text{CE}}\left(\hat{\mathbf{y}}(x), y\right) \right]
\end{equation}

Although we can get predictions via neural symbolic layer, we can also see such prediction is purely based on the concept rules we have learned without fully leveraging the feature vectors, indicating that directly using such a classifier will have bad performance. In the final step of our framework, we will improve the final performance.

\subsection{Final Prediction and Network Optimization}

The overall training objective consists of multiple components: standard classification accuracy, concept learning, and explainability constraints. The loss function is formulated as:

\begin{equation}
\mathcal{L} = \mathcal{L}_{prediction} + \alpha_1 \mathcal{L}_{concept} + \alpha_2 \mathcal{L}_{nueral}
\end{equation}


\noindent where $\mathcal{L}_{prediction}=\mathbb{E}[\ell_{CE}(y^*(x), y)]$, $\mathcal{L}_{concept}=\mathbb{E}[\ell_{CE}(\mathcal{C}^{*}, \mathcal{C})]$ and $\mathcal{L}_{nueral}=\mathbb{E}[\ell_{CE}(\mathrm{y}^*(x), y)]$.

\section{Experiments}
\subsection{Experimental Settings}
\subsubsection{Datasets}

The dataset utilized in this study is derived from the CEBaB dataset, focusing on sentiment classification for restaurant reviews. Specifically, we used the augmented version of the dataset, referred to as \textbf{aug-CEBaB-yelp}. This dataset consists of two components: a source concept dataset ($D_s$) containing human-annotated concepts, such as \textit{Food}, \textit{Ambiance}, \textit{Service}, and \textit{Noise}, and an unlabeled concept dataset ($D_u$) derived from Yelp reviews, providing a large number of samples for augmentation. Following augmentation, the dataset is transformed into $\tilde{D} = \{\tilde{D}_{sa}, \tilde{D}_u\}$, where $\tilde{D}_{sa}$ combines human-annotated concepts with augmented concepts having noisy labels, and $\tilde{D}_u$ includes automatically labeled data containing both human-specified concepts and ChatGPT-generated concepts. Each concept in the dataset takes one of three possible values: \textit{Positive}, \textit{Negative}, or \textit{Unknown}.

To enrich the dataset, we applied a data augmentation process consisting of concept set augmentation and noisy concept label annotation. For concept set augmentation, the goal was to generate additional high-quality concepts to expand the original concept set ($C_s$). Using ChatGPT and employing an in-context learning strategy, we provided human-specified concepts as references to guide the generation of new concepts. For example, the prompt, “Besides \{Food, Ambiance, Service, Noise\}, what are the additional important features to evaluate a restaurant review?” was used, where \textit{Food}, \textit{Ambiance}, \textit{Service}, and \textit{Noise} are manually annotated concepts from $C_s$. ChatGPT-generated concepts, such as \textit{Cleanliness} and \textit{Menu Variety}, were then filtered to remove irrelevant or rare outputs, resulting in an augmented concept set ($C_a$). 

In the noisy concept label annotation step, ChatGPT was employed to annotate unlabeled data with noisy labels for both human-specified and ChatGPT-generated concepts. This was achieved by designing prompts that provided human-annotated examples as context. For instance, a sample prompt is: “a. According to the review \{text1\}, the \{concept1\} of the restaurant is \textit{positive}. b. According to the review \{text2\}, the \{concept2\} of the restaurant is \textit{negative}. c. According to the review \{text3\}, the \{concept3\} of the restaurant is \textit{unknown}. d. According to the review \{texti\}, how is the \{concepti\} of the restaurant? Please answer with one option in \textit{positive}, \textit{negative}, or \textit{unknown}.” Using such prompts, ChatGPT annotated the concepts for unlabeled reviews, including both original concepts from $D_s$ and new concepts from $C_a$. The resulting $\tilde{D}_u$ dataset combines noisy labels for both human-specified and ChatGPT-generated concepts, enhancing the diversity and scope of the data. The augmented dataset $\tilde{D}$ integrates high-quality human annotations with a comprehensive set of ChatGPT-generated concepts and noisy labels, providing a robust foundation for training and evaluating sentiment classification models for restaurant reviews.

\subsubsection{Backbones}

Four distinct language models were adopted as backbones for our study.

\noindent\textbf{Bert-base-uncased}, developed by Google, is a bidirectional Transformer model. With 12 layers, 768 hidden units and 12 attention heads, it extracts context from both directions, leveraging pre-trained knowledge on diverse corpora for sentiment understanding.

\noindent\textbf{Roberta-base} is an optimized variant of BERT. Despite sharing a similar architecture, it's trained with larger batch sizes and longer schedules, enabling better generalization to our dataset.

\noindent\textbf{Gpt2}, developed by OpenAI, is an autoregressive model. Different from BERT-like models, it generates text token-by-token. The base version, with 12 layers, 768 hidden units and 12 attention heads, can capture complex semantic dependencies.

\noindent\textbf{Lstm}, a type of recurrent neural network, processes sequential data. Its memory cell structure allows it to capture long-term dependencies in review text, learning sentiment-related patterns during training on our dataset. 

These backbones’ performance with and without the CLMN structure helps assess the effectiveness of our approach.

\subsubsection{Metrics}

To evaluate both the utility and interpretability of our models, we use the following metrics:

\begin{table*}[th]
    \centering
    \caption{Performance comparison between backbone models and their corresponding CLMN outputs.}
    \label{tab:model_performance}
    \begin{tabular}{lcc|cccccc}
        \hline
        & \multicolumn{2}{c|}{Backbone} & \multicolumn{6}{c}{CLMN} \\
        Model & Acc & F1 & O-Acc & O-F1 & C-Acc & C-F1 & R-Acc & R-F1 \\
        \hline
        bert-base-uncased & 69.49 & 79.72 & 69.26 & 79.62 & 85.85 & 84.63 & \textbf{65.35} & 76.49 \\
        roberta-base & \textbf{71.21} & \textbf{80.92} & \textbf{71.56} & \textbf{81.16} & \textbf{87.52} & \textbf{86.09} & 64.68 & \textbf{76.51} \\
        gpt2 & 63.39 & 75.39 & 63.60 & 76.39 & 87.12 & 85.18 & 62.72 & 75.76 \\
        lstm & 47.54 & 65.65 & 47.19 & 64.03 & 75.65 & 66.60 & 38.08 & 57.10 \\
        \hline
    \end{tabular}
    \vspace{0.5em}
    
    \small
    \textit{O = Output-level (final prediction), C = Concept-level, R = Reasoning-level. Acc = Accuracy, F1 = Macro F1.}
\end{table*}

\begin{table*}[th]
    \centering
    \caption{Ablation study on concept\_weight and y2\_weight parameters.}
    \begin{tabular}{lcccccc}
        \hline
        Model & Acc & Macro F1 & Concept Acc & Concept Macro F1 & Test Acc & Test Macro F1 \\
        \hline
        concept\_weight=0, y2\_weight=0 & 68.63 & 78.90 & 71.76 & 44.61 & 8.35 & 39.52  \\
        concept\_weight=100, y2\_weight=0 & 69.57 & 79.63 & 87.26 & 85.63 & 19.59 & 47.95  \\
        concept\_weight=0, y2\_weight=10 & 67.51 & 79.00 & 26.46 & 39.90 & 67.32 & 78.85  \\
        concept\_weight=100, y2\_weight=10 & 69.26 & 79.62 & 85.85 & 84.63 & 65.35 & 76.49  \\
        \hline
    \end{tabular}
    \label{tab:ablation_study}
\end{table*}

\textbf{Task Performance Metrics:}
\begin{itemize}
    \item \textbf{Task Accuracy:} This metric measures the proportion of correctly predicted task labels, providing a straightforward evaluation of classification performance.
    \item \textbf{Task Macro F1 Score:} The macro F1 score is calculated as the harmonic mean of precision and recall in all classes, equally weighted regardless of class size. This metric is particularly suitable for datasets with imbalanced class distributions.
\end{itemize}

\textbf{Interpretability Metrics:}
\begin{itemize}
    \item \textbf{Concept Accuracy:} This metric evaluates the model's ability to accurately predict concept labels, reflecting the interpretability of the learned representations.
    \item \textbf{Concept Macro F1 Score:} Similar to the task-level F1 score, this metric measures the harmonic mean of Precision and Recall at the concept level, offering a balanced evaluation of interpretability performance across all concepts.
\end{itemize}

These metrics are used to assess the trade-off between interpretability and utility across various datasets and training strategies.

\subsubsection{Experimental Setup}

In our experiment, we conducted parameter experiments to evaluate the impact of different parameter settings on the model's performance, as well as ablation studies to analyze the contributions of individual components or modules to the overall effectiveness of the approach. Through several rounds of parameter tuning and comparative analysis, we set the maximum token number of input to 512, the number of training epochs to 25, and the batch size to 8. The weights for the concept loss and neural loss were set as $\alpha_1 = 100$ and $\alpha_2 = 10$, respectively. The learning rates were configured individually for different backbone models: $1\mathrm{e}{-5}$ for BERT-base and RoBERTa-base, $1\mathrm{e}{-4}$ for GPT-2, and $1\mathrm{e}{-2}$ for the LSTM-based architecture.

\subsection{Model Utility AND Interpretability}

\begin{figure}[htb]
\centering 
\fbox{\includegraphics[width=0.98\linewidth]{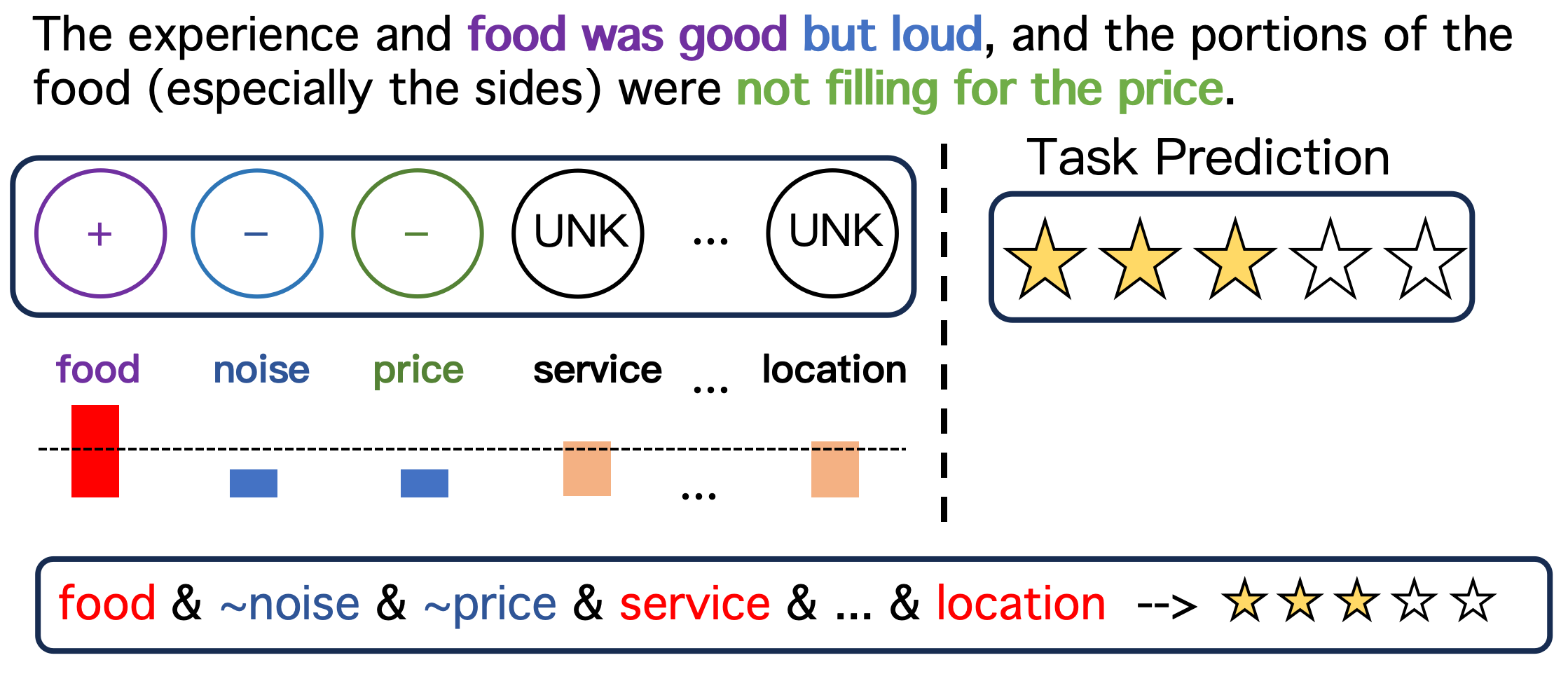}}
\caption{Interpretability Example}
\label{fig:interprebility-example}
\end{figure}

According to Table \ref{tab:model_performance}, the BERT model integrated with the CLMN framework achieves concept prediction accuracy ($\text{Acc}$) and $\text{Macro F1}$ scores of $\mathbf{85.85\%}$ and $\mathbf{84.63\%}$ respectively, demonstrating effective concept recognition capabilities. For final task prediction, the reasoning network attains $\mathbf{65.35\%}$ $\text{Acc}$ and $\mathbf{76.49\%}$ $\text{Macro F1}$ through concept-based inference, indicating the model's capacity to leverage intermediate concepts for interpretable decision-making rather than relying solely on black-box predictions. The $\mathbf{3.91\%}$ accuracy gap between direct prediction ($69.26\%$ $\text{Acc}$) and concept-driven reasoning suggests certain final labels cannot be fully explained through the identified concepts. By strategically employing direct prediction for final outputs while retaining concept reasoning traces, the model maintains competitive performance ($69.26\%$ $\text{Acc}$, $79.62\%$ $\text{Macro F1}$) comparable to non-interpretable backbones, while providing explicit derivation pathways for most cases. This architecture successfully preserves classification accuracy within $\mathbf{0.3\%}$ relative degradation compared to baseline models, proving that interpretability requirements can be incorporated almost without compromising final task performance. The residual reasoning gap highlights potential directions for strengthening concept-task alignment in future work.

\subsection{Ablation Study}

To discuss the functions of each component in more detail, we conducted a series of comprehensive ablation experiments. Our findings reveal that the introduction of the concept loss barely weakens the performance of the model on the final task. In contrast, the incorporation of only the neural symbolic component leads to a certain degree of reduction in the final prediction ability of the model. This is primarily because the additional task effectively diverts the attention allocation during the model's training process. Without the control of concept prediction, the consistency between the two types of predictions is relatively weak.
However, when both the concept loss component and the neural symbolic component are added simultaneously, the model's final prediction ability is almost on par with that of the baseline model. Moreover, it maintains the correctness of the reasoning rules of the neural symbolic component, significantly enhancing the interpretability of the entire model. Detailed results of the ablation study are presented in Table \ref{tab:ablation_study}.

\subsection{Interpretability}
The example in Figure~\ref{fig:interprebility-example} demonstrates the interpretability visualization process of the CLMN framework. In the original sentence, phrases such as "\textit{food was good}", "\textit{loud}", and "\textit{not filling for the price}" provide evidence for concept predictions corresponding to \textit{food}, \textit{noise}, and \textit{price} respectively. The \textit{Unknown} labels in the concept layer, while potentially participating in subsequent neural symbolic processes, exhibit less significant contribution tendency compared to definitive concepts like \textit{food} and \textit{loud} during rule formation. The construction of symbolic rules requires comprehensive consideration of the entire training dataset to ensure generalization capability.

\section{Conclusion}

This work presents CLMN, a neural-symbolic framework that bridges performance and interpretability in NLP through continuous concept embeddings and fuzzy logic operations. By replacing traditional binary concept activations with learnable neural-symbolic operators, CLMN preserves contextual semantics while enabling dynamic concept interactions and transparent rule extraction. Experiments across multiple models and datasets show CLMN achieves competitive accuracy with superior concept alignment, demonstrating that interpretability need not compromise performance. The framework’s decoupled prediction-explainability architecture offers a viable path for deploying trustworthy AI in healthcare and finance, with future extensions targeting hierarchical concept reasoning and multilingual adaptability.

\bibliography{main}
\bibliographystyle{abbrv}
\addtolength{\textheight}{-12cm}   




\end{document}